\documentclass[conference]{IEEEtran}
\IEEEoverridecommandlockouts
\usepackage{amsmath,amssymb,amsfonts}
\usepackage{algorithmic}
\usepackage{hyperref}
\usepackage{graphicx}
\usepackage{textcomp}
\usepackage{xcolor}
\usepackage{acronym}
\usepackage{float}
\let\mb\mathbf
\usepackage[backend=biber, sorting=none, style=numeric-comp, maxnames=5]{biblatex}

\acrodef{AC}[AC]{Coefficient of Variation}
\acrodef{ADC}[ADC]{Analog-to-Digital Converter}
\acrodef{ADEXP}[AdExp-I\&F]{Adaptive-Exponential Integrate and Fire}
\acrodef{ADM}[ADM]{Asynchronous Delta Modulator}
\acrodef{AER}[AER]{Address-Event Representation}
\acrodef{AEX}[AEX]{AER EXtension board}
\acrodef{AE}[AE]{Address-Event}
\acrodef{AFM}[AFM]{Atomic Force Microscope}
\acrodef{AGC}[AGC]{Automatic Gain Control}
\acrodef{AI}[AI]{Artificial Intelligence}
\acrodef{AMDA}[AMDA]{AER Motherboard with D/A converters}
\acrodef{AMPA}[AMPA]{$\alpha$-amino-3-hydroxy-5-methyl-4-isoxazolepropionic acid}
\acrodef{ANN}[ANN]{Artificial Neural Network}
\acrodef{API}[API]{Application Programming Interface}
\acrodef{APMOM}[APMOM]{Alternate Polarity Metal On Metal}
\acrodef{ARM}[ARM]{Advanced RISC Machine}
\acrodef{ASIC}[ASIC]{Application Specific Integrated Circuit}
\acrodef{AdExp}[AdExp-IF]{Adaptive Exponential Integrate-and-Fire}
\acrodef{BCM}[BMC]{Bienenstock-Cooper-Munro}
\acrodef{BD}[BD]{Bundled Data}
\acrodef{BEOL}[BEOL]{Back-end of Line}
\acrodef{BG}[BG]{Bias Generator}
\acrodef{BMI}[BMI]{Brain-Machince Interface}
\acrodef{BTB}[BTB]{band-to-band tunnelling}
\acrodef{CAD}[CAD]{Computer Aided Design}
\acrodef{CAM}[CAM]{Content Addressable Memory}
\acrodef{CAVIAR}[CAVIAR]{Convolution AER Vision Architecture for Real-Time}
\acrodef{CA}[CA]{Cortical Automaton}
\acrodef{CCN}[CCN]{Cooperative and Competitive Network}
\acrodef{CDR}[CDR]{Clock-Data Recovery}
\acrodef{CFC}[CFC]{Current to Frequency Converter}
\acrodef{CHP}[CHP]{Communicating Hardware Processes}
\acrodef{CMIM}[CMIM]{Metal-insulator-metal Capacitor}
\acrodef{CML}[CML]{Current Mode Logic}
\acrodef{CMOL}[CMOL]{Hybrid CMOS nanoelectronic circuits}
\acrodef{CMOS}[CMOS]{complementary metal–oxide–semiconductor}
\acrodef{CNN}[CNN]{Convolutional Neural Network}
\acrodef{CNS}[CNS]{central nervous system}
\acrodef{COTS}[COTS]{Commercial Off-The-Shelf}
\acrodef{CPG}[CPG]{Central Pattern Generator}
\acrodef{CPLD}[CPLD]{Complex Programmable Logic Device}
\acrodef{CPU}[CPU]{Central Processing Unit}
\acrodef{CSM}[CSM]{Cortical State Machine}
\acrodef{CSP}[CSP]{Constraint Satisfaction Problem}
\acrodef{CTXCTL}[CTXCTL]{CortexControl}
\acrodef{CV}[CV]{Coefficient of Variation}
\acrodef{DAC}[DAC]{Digital to Analog Converter}
\acrodef{DAS}[DAS]{Dynamic Auditory Sensor}
\acrodef{DAVIS}[DAVIS]{Dynamic and Active Pixel Vision Sensor}
\acrodef{DBN}[DBN]{Deep Belief Network}
\acrodef{DBS}[DBS]{Deep Brain Stimulation}
\acrodef{DFA}[DFA]{Deterministic Finite Automaton}
\acrodef{DIBL}[DIBL]{drain-induced-barrier-lowering}
\acrodef{DI}[DI]{delay insensitive}
\acrodef{DMA}[DMA]{Direct Memory Access}
\acrodef{DNF}[DNF]{Dynamic Neural Field}
\acrodef{DNN}[DNN]{Deep Neural Network}
\acrodef{DOF}[DOF]{Degrees of Freedom}
\acrodef{DPE}[DPE]{Dynamic Parameter Estimation}
\acrodef{DPI}[DPI]{Differential Pair Integrator}
\acrodef{DRAM}[DRAM]{Dynamic Random Access Memory}
\acrodef{DRRZ}[DR-RZ]{Dual-Rail Return-to-Zero}
\acrodef{DR}[DR]{Dual Rail}
\acrodef{DSP}[DSP]{Digital Signal Processor}
\acrodef{DVS}[DVS]{Dynamic Vision Sensor}
\acrodef{DYNAP}[DYNAP]{Dynamic Neuromorphic Asynchronous Processor}
\acrodef{EBL}[EBL]{Electron Beam Lithography}
\acrodef{ECG}[ECG]{Electrocardiography}
\acrodef{ECoG}[ECoG]{Electrocorticography}
\acrodef{EDVAC}[EDVAC]{Electronic Discrete Variable Automatic Computer}
\acrodef{EEG}[EEG]{Electroencephalography}
\acrodef{EIN}[EIN]{Excitatory-Inhibitory Network}
\acrodef{EMG}[EMG]{Electromyography}
\acrodef{EM}[EM]{Expectation Maximization}
\acrodef{EOG}[EOG]{Electrooculogram}
\acrodef{EPSC}[EPSC]{Excitatory Post-Synaptic Current}
\acrodef{EPSP}[EPSP]{Excitatory Post-Synaptic Potential}
\acrodef{EZ}[EZ]{Epileptogenic Zone}
\acrodef{FDSOI}[FDSOI]{Fully-Depleted Silicon on Insulator}
\acrodef{FET}[FET]{Field-Effect Transistor}
\acrodef{FFT}[FFT]{Fast Fourier Transform}
\acrodef{FI}[F-I]{Frequency-Current}
\acrodef{FMA}[FMA]{Floating microelectrode array} 
\acrodef{FNN}[FNN]{Feed-forward Neural Network}
\acrodef{FPGA}[FPGA]{Field Programmable Gate Array}
\acrodef{FR}[FR]{Fast Ripple}
\acrodef{FSA}[FSA]{Finite State Automaton}
\acrodef{FSM}[FSM]{Finite State Machine}
\acrodef{GABA}[GABA]{$\gamma$-aminobutanoic acid}
\acrodef{GIDL}[GIDL]{gate-induced-drain-leakage}
\acrodef{GOPS}[GOPS]{Giga-Operations per Second}
\acrodef{GPU}[GPU]{Graphical Processing Unit}
\acrodef{GT}[GT]{Ground Truth}
\acrodef{GUI}[GUI]{Graphical User Interface}
\acrodef{HAL}[HAL]{Hardware Abstraction Layer}
\acrodef{HFO}[HFO]{High Frequency Oscillation}
\acrodef{HH}[H\&H]{Hodgkin \& Huxley}
\acrodef{HMM}[HMM]{Hidden Markov Model}
\acrodef{HRS}[HRS]{High-Resistive State}
\acrodef{HR}[HR]{Human Readable}
\acrodef{HSE}[HSE]{Handshaking Expansion}
\acrodef{HW}[HW]{Hardware}
\acrodef{ICT}[ICT]{Information and Communication Technology}
\acrodef{IC}[IC]{Integrated Circuit}
\acrodef{IF2DWTA}[IF2DWTA]{Integrate \& Fire 2--Dimensional WTA}
\acrodef{IFSLWTA}[IFSLWTA]{Integrate \& Fire Stop Learning WTA}
\acrodef{IF}[I\&F]{Integrate-and-Fire}
\acrodef{IMU}[IMU]{Inertial Measurement Unit}
\acrodef{INCF}[INCF]{International Neuroinformatics Coordinating Facility}
\acrodef{INI}[INI]{Institute of Neuroinformatics}
\acrodef{IO}[I/O]{Input/Output}
\acrodef{IPSC}[IPSC]{Inhibitory Post-Synaptic Current}
\acrodef{IPSP}[IPSP]{Inhibitory Post-Synaptic Potential}
\acrodef{IP}[IP]{Intellectual Property}
\acrodef{ISI}[ISI]{Inter-Spike Interval}
\acrodef{IoT}[IoT]{Internet of Things}
\acrodef{JFLAP}[JFLAP]{Java - Formal Languages and Automata Package}
\acrodef{LEDR}[LEDR]{Level-Encoded Dual-Rail}
\acrodef{LFP}[LFP]{Local Field Potential}
\acrodef{LIFE}[LIFE]{Longitudinal Intrafascicular Electrodes}
\acrodef{LIF}[LI\&F]{Leak Integrate-and-Fire}
\acrodef{LLC}[LLC]{Low Leakage Cell}
\acrodef{LNA}[LNA]{Low-Noise Amplifier}
\acrodef{LPF}[LPF]{Low Pass Filter}
\acrodef{LRS}[LRS]{Low-Resistive State}
\acrodef{LSM}[LSM]{Liquid State Machine}
\acrodef{LTD}[LTD]{Long Term Depression}
\acrodef{LTI}[LTI]{Linear Time-Invariant}
\acrodef{LTP}[LTP]{Long Term Potentiation}
\acrodef{LTU}[LTU]{Linear Threshold Unit}
\acrodef{LUT}[LUT]{Look-Up Table}
\acrodef{LVDS}[LVDS]{Low Voltage Differential Signaling}
\acrodef{MCMC}[MCMC]{Markov-Chain Monte Carlo}
\acrodef{MEA}[MEA]{Multielectrode Arrays}
\acrodef{MEMS}[MEMS]{Micro Electro Mechanical System}
\acrodef{MFR}[MFR]{Mean Firing Rate}
\acrodef{MIM}[MIM]{Metal Insulator Metal}
\acrodef{MLP}[MLP]{Multilayer Perceptron}
\acrodef{ML}[ML]{Machine Learning}
\acrodef{MOSCAP}[MOSCAP]{Metal Oxide Semiconductor Capacitor}
\acrodef{MOSFET}[MOSFET]{Metal Oxide Semiconductor Field-Effect Transistor}
\acrodef{MOS}[MOS]{Metal Oxide Semiconductor}
\acrodef{MRI}[MRI]{Magnetic Resonance Imaging}
\acrodef{NCS}[NCS]{Neuromorphic Cognitive Systems}
\acrodef{NDFSM}[NDFSM]{Non-deterministic Finite State Machine} 
\acrodef{ND}[ND]{Noise-Driven}
\acrodef{NEF}[NEF]{Neural Engineering Framework}
\acrodef{NHML}[NHML]{Neuromorphic Hardware Mark-up Language}
\acrodef{NIL}[NIL]{Nano-Imprint Lithography}
\acrodef{NI}[NI]{Neural Interface}
\acrodef{NMDA}[NMDA]{N-Methyl-D-Aspartate}
\acrodef{NME}[NE]{Neuromorphic Engineering}
\acrodef{NN}[NN]{Neural Network}
\acrodef{NOC}[NoC]{Network-on-Chip}
\acrodef{NRZ}[NRZ]{Non-Return-to-Zero}
\acrodef{NSM}[NSM]{Neural State Machine}
\acrodef{OR}[OR]{Operating Room}
\acrodef{OTA}[OTA]{Operational Transconductance Amplifier}
\acrodef{PCB}[PCB]{Printed Circuit Board}
\acrodef{PCHB}[PCHB]{Pre-Charge Half-Buffer}
\acrodef{PCM}[PCM]{Phase Change Memory}
\acrodef{PCA}[PCA]{Personal Component Analysis}

\acrodef{PC}[PC]{Personal Computer}
\acrodef{PE}[PE]{Phase Encoding}
\acrodef{PFA}[PFA]{Probabilistic Finite Automaton}
\acrodef{PFC}[PFC]{prefrontal cortex}
\acrodef{PFM}[PFM]{Pulse Frequency Modulation}
\acrodef{PNI}[PNI]{peripheral nerve interface}
\acrodef{PNS}[PNS]{peripheral nervous system}
\acrodef{PPG}[PPG]{Photoplethysmography}
\acrodef{PR}[PR]{Production Rule}
\acrodef{PSC}[PSC]{Post-Synaptic Current}
\acrodef{PSP}[PSP]{Post-Synaptic Potential}
\acrodef{PSTH}[PSTH]{Peri-Stimulus Time Histogram}
\acrodef{PV}[PV]{Parvalbumin}
\acrodef{QDI}[QDI]{Quasi Delay Insensitive}
\acrodef{RAM}[RAM]{Random Access Memory}
\acrodef{RA}[RA]{Resected Area}
\acrodef{RDF}[RDF]{random dopant fluctuation}
\acrodef{RELU}[ReLu]{Rectified Linear Unit}
\acrodef{RLS}[RLS]{Recursive Least-Squares}
\acrodef{RMSE}[RMSE]{Root Mean Squared-Error}
\acrodef{RMS}[RMS]{Root Mean Squared}
\acrodef{RNN}[RNN]{Recurrent Neural Networks}
\acrodef{RNN}[RNN]{Recurrent Neural Network}
\acrodef{ROLLS}[ROLLS]{Reconfigurable On-Line Learning Spiking}
\acrodef{RRAM}[R-RAM]{Resistive Random Access Memory}
\acrodef{RSA}[RSA]{Respiratory Sinus Arrhythmia}
\acrodef{R}[R]{Ripples}
\acrodef{SAC}[SAC]{Selective Attention Chip}
\acrodef{SAT}[SAT]{Boolean Satisfiability Problem}
\acrodef{SCI}[SCI]{Spinal Cord Injury}
\acrodef{SCX}[SCX]{Silicon CorteX}
\acrodef{SD}[SD]{Signal-Driven}
\acrodef{SEM}[SEM]{Spike-based Expectation Maximization}
\acrodef{SLAM}[SLAM]{Simultaneous Localization and Mapping}
\acrodef{SNN}[SNN]{Spiking Neural Network}
\acrodef{SNR}[SNR]{Signal to Noise Ratio}
\acrodef{SOC}[SOC]{System-On-Chip}
\acrodef{SOI}[SOI]{Silicon on Insulator}
\acrodef{SOZ}[SOZ]{Seizure Onset Zone}
\acrodef{SP}[SP]{Separation Property}
\acrodef{SRAM}[SRAM]{Static Random Access Memory}
\acrodef{PYR}[PYR]{Pyramidal}
\acrodef{SST}[SST]{Somatostatin}

\acrodef{STDP}[STDP]{Spike-Timing Dependent Plasticity}
\acrodef{STD}[STD]{Short-Term Depression}
\acrodef{STP}[STP]{Short-Term Plasticity}
\acrodef{STT-MRAM}[STT-MRAM]{Spin-Transfer Torque Magnetic Random Access Memory}
\acrodef{STT}[STT]{Spin-Transfer Torque}
\acrodef{SVM}[SVM]{Support Vector Machine}
\acrodef{SW}[SW]{Software}
\acrodef{TCAM}[TCAM]{Ternary Content-Addressable Memory}
\acrodef{TFT}[TFT]{Thin Film Transistor}
\acrodef{TIME}[TIME]{Transverse Intrafascicular Multichannel Electrode}
\acrodef{TLE}[TLE]{Temporal Lobe Epilepsy}
\acrodef{UEA}[UEA]{Utah electrode array}
\acrodef{USB}[USB]{Universal Serial Bus}
\acrodef{USEA}[USEA]{Utah Slanted Electrode Array}
\acrodef{VHDL}[VHDL]{VHSIC Hardware Description Language}
\acrodef{VIP}[VIP]{Vasoactive Intestinal Peptide}
\acrodef{VLSI}[VLSI]{Very Large Scale Integration}
\acrodef{VNS}[VNS]{Vagal Nerve Stimulation}
\acrodef{VOR}[VOR]{Vestibulo-Ocular Reflex}
\acrodef{WCST}[WCST]{Wisconsin Card Sorting Test}
\acrodef{WTA}[WTA]{Winner-Take-All}
\acrodef{XML}[XML]{eXtensible Mark-up Language}
\acrodef{divmod3}[DIVMOD3]{divisibility of a number by three}
\acrodef{hWTA}[hWTA]{hard Winner-Take-All}
\acrodef{iEEG}[iEEG]{intracranial electroencephalography}
\acrodef{sWTA}[sWTA]{soft Winner-Take-All}

\DeclareSourcemap{
  \maps[datatype=bibtex]{
    \map[overwrite=true]{
      \step[fieldset=url, null]
      \step[fieldset=eprint, null]
    }
  }
}

\newcommand{\appropto}{\mathrel{\vcenter{
  \offinterlineskip\halign{\hfil$##$\cr
    \propto\cr\noalign{\kern2pt}\sim\cr\noalign{\kern-2pt}}}}}
    
\def\BibTeX{{\rm B\kern-.05em{\sc i\kern-.025em b}\kern-.08em
    T\kern-.1667em\lower.7ex\hbox{E}\kern-.125emX}}
\begin{document}

\title{A feedback control optimizer for online and hardware-aware training of Spiking Neural Networks}

\author{\IEEEauthorblockN{Matteo Saponati\IEEEauthorrefmark{1},
Chiara De Luca\IEEEauthorrefmark{1}\IEEEauthorrefmark{2},
Giacomo Indiveri\IEEEauthorrefmark{1},~\IEEEmembership{Senior Member, IEEE}, and 
Benjamin Grewe\IEEEauthorrefmark{1}\IEEEauthorrefmark{3}}
\IEEEauthorblockA{\IEEEauthorrefmark{1}Institute of Neuroinformatics, University of Zürich and ETH Zürich, Zürich, Switzerland}
\IEEEauthorblockA{\IEEEauthorrefmark{2}Digital Society Initiative, University of Zurich, Switzerland}
\IEEEauthorblockA{\IEEEauthorrefmark{3}AI Center, ETH, Zürich, Switzerland}
\thanks{}}

\maketitle

\begin{abstract}
Unlike traditional artificial neural networks (ANNs), biological neuronal networks solve complex cognitive tasks with sparse neuronal activity, recurrent connections, and local learning rules.
These mechanisms serve as design principles in Neuromorphic computing, which addresses the critical challenge of energy consumption in modern computing.
However, most mixed-signal neuromorphic devices rely on semi- or unsupervised learning rules, which are ineffective for optimizing hardware in supervised learning tasks.
This lack of scalable solutions for on-chip learning restricts the potential of mixed-signal devices to enable sustainable, intelligent edge systems.
To address these challenges, we present a novel learning algorithm for Spiking Neural Networks (SNNs) on mixed-signal devices that integrates spike-based weight updates with feedback control signals. 
In our framework, a spiking controller generates feedback signals to guide SNN activity and drive weight updates, enabling scalable and local on-chip learning.
We first evaluate the algorithm on various classification tasks, demonstrating that single-layer SNNs trained with feedback control achieve performance comparable to artificial neural networks (ANNs).
We then assess its implementation on mixed-signal neuromorphic devices by testing network performance in continuous online learning scenarios and evaluating resilience to hyperparameter mismatches.
Our results show that the feedback control optimizer is compatible with neuromorphic applications, advancing the potential for scalable, on-chip learning solutions in edge applications.
\end{abstract}


\section{Introduction}
\label{sec:introduction}
Nowadays, deep ANNs have become a cornerstone of modern artificial intelligence \cite{lecun2015deep}.
This has been driven by advancements in hardware acceleration (i.e., GPUs, TPUs, and NPUs) and intensive training procedures with backpropagation of error \cite{rumelhartLearningRepresentationsBackpropagating1986} over hundreds to thousands of computing machines.
While backpropagation allows for massive parallelization of training, it also requires intensive memory access by
(1) storing and retrieving intermediate state variables for automatic differentiation, (2) encoding error information with separate states, and (3) updating every parameter during every backward pass ~\cite{strubell2020energy,lillicrapBackpropagationBrain2020}.
This approach makes ANN ideal for tasks like image classification, speech recognition, and large-scale language modeling~\cite{Plis_deep2014, Mnih_human2015, vaswani2017attention}, where massive memory access and training on large, pre-collected datasets in controlled settings are possible (offline learning).
On the other hand, these models are unsuitable for applications where power efficiency, continuous, and adaptive optimization are key requirements (online learning) \cite{chen2019deep}.
This online learning approach is particularly critical for systems operating in an always-on mode, enabling them to adapt continuously to changing environments in edge applications such as autonomous systems, IoT devices, and personalized services.

A change of paradigm is thus needed to address such challenges. The field of neuromorphic computing has embraced this shift by drawing inspiration from the brain's remarkable ability to learn continuously, efficiently, and in a distributed manner~\cite{roy2019towards}. 
Biological neural networks operate with local learning mechanisms that are adaptive, scalable, and optimized for real-time functionality~\cite{indiveri2015memory, marblestone2016toward}.
These networks use spike-based communication and feedback loops, enabling efficient, on-the-fly learning without the need for massive data storage or global error propagation \cite{roy2019towards}.
The challenge of designing systems that meet the power and adaptability needs of edge devices calls for more than just energy-efficient hardware; it requires brain-inspired algorithms that fully exploit the capabilities of neuromorphic hardware. 

However, current learning algorithms for mixed-signal devices face significant limitations in both performance and expressivity. 
Indeed, bio-inspired learning in neuromorphic hardware has been limited to unsupervised rules \cite{izhikevich2007solving, gerstner1996stdp, pfisterTripletsSpikesModel2006,clopathVoltageSpikeTiming2010,krotov2019unsupervised, patel2020unsupervised}, which often require extensive hyperparameter tuning and the integration of multiple hard-coded stabilization mechanisms to function effectively \cite{masquelierSpikeTimingDependent2008, taherkhani2020review}.
Furthermore, there is limited theoretical understanding of how to extend these mechanisms to multi-layered spiking neural networks (SNNs) \cite{pfeifferDeepLearningSpiking2018}.
Moreover, computer in-the-loop (ITL) methods with surrogate gradients have shown promise for training multi-layer SNNs on mixed-signal devices\cite{cramerSurrogateGradientsAnalog2022}.
However, ITL approaches require storing the computation graph of the SNN, assuming idealized dynamics, computing the weight updates with backpropagation through time (BPTT), and reconfiguring the network at every training step.
This process is energy-intensive and unsuitable for online learning in adaptable devices where energy efficiency is critical.
As yet, training SNNs online and directly on neuromorphic devices remains a challenging engineering and computational problem.

In this work, we address this challenge by proposing a novel optimization algorithm that integrates spike-based local learning rules with feedback control signals.
Our approach is inspired by recent work on biologically plausible credit assignment with control \cite{sacramentoDendriticCorticalMicrocircuits2018,meulemansCreditAssignmentNeural2021, meulemansMinimizingControlCredit2022,meulemansLeastcontrolPrincipleLocal2022,aceitunoLearningCorticalHierarchies2023}. 
In this family of models, a control module is used to steer network activities toward target outputs and generate learning signals for effective credit assignment in deep ANNs \cite{meulemansCreditAssignmentNeural2021}.
Here, we implement these mechanisms in an SNN, introducing a novel architecture with a control module of spiking neurons.
In our model, both inference and learning are mediated by the recurrent connections between these two modules.
The error information is encoded in the temporal dynamics of synaptic currents, eliminating the need to store intermediate network states.
This enables neurons to adjust synaptic weights online, removing the requirement for separate training and inference phases.
Additionally,  weight updates follow a fully local learning rule guided by the control signals, without relying on nested gradient computations typical of backpropagation.

First, we train SNNs on multiple supervised learning tasks with our feedback control optimizer. 
Second, we demonstrate that our SNN feedback control algorithm is suitable for online learning, and we validate its performance by emulating typical hardware constraints (i.e., number of connections and hardware substrate mismatch). 
In conclusion, this study demonstrates the potentiation of this novel hardware-aware algorithm to revolutionize the scalability of neuromorphic devices.

\section{The feedback control optimizer}
\subsection{The network architecture}
\begin{figure*}[htbp]
\includegraphics[width=\textwidth]{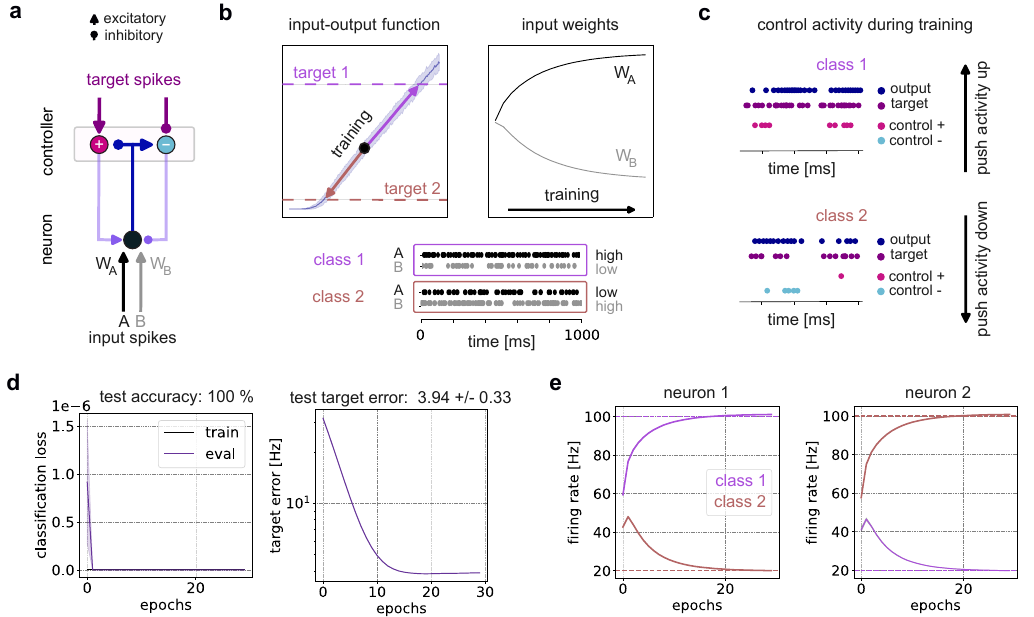}
\caption{\textbf{Illustration of the feedback control optimizer and binary classification task}.
\textbf{a}) Illustration of the feedback control architecture.
A neuron receives external input spikes (black and grey arrows) associated with learnable synaptic weights ($W_A$ and $W_B$) and, in turn,  sends output spikes to the control module (blue arrows).
The neuron is paired with a positive (magenta) and a negative (light blue) control neuron.
The positive and negative control neurons receive inhibitory and excitatory activity from the output neuron, respectively.
The control neurons also receive external target spikes (purple arrows) and, in turn, send feedback spikes (light purple arrows).
\textbf{b}) Illustration of the binary classification task.
Bottom) The activity of the pre-synaptic inputs A and B  for class 1 (pink) and 2 (orange), respectively.
Top-left) The input-output function (f-f curve) of the neuron.
The neuron has two target activities corresponding to the two classes (target 1 and target 2) and is trained to match these targets when inputs from the respective classes are presented.
Top-right) Evolution of the learnable synaptic weights $W_A$ and $W_B$ during training.
\textbf{c}) Illustration of the feedback control algorithm.
Example spiking activities of the neuron, the targets, and the positive and negative control for an example input from class 1 (top) and class 2 (bottom).
\textbf{d})
Left: the average cross-entropy loss during training (black) and validation (purple).
Right: the mean absolute difference between output and target activity during validation.
\textbf{e})
Average firing rate during validation for inputs from class 1 (black) and class 2 (light red). 
Left: neuron encoding class 1. Right: neuron encoding class 2.
We calculate the mean and standard deviation across multiple simulations with different random seeds.
}
\label{fig:model-binary-classification}
\end{figure*}
%
We built the architecture of the feedback control optimizer on a simple computational primitive: (1) an Integrate-and-Fire (LIF) output neuron, and (2) a controller module of one excitatory (positive) and one inhibitory (negative) LIF control neuron (Figure \ref{fig:model-binary-classification}a, left). 
During training, the neuron receives external inputs (bottom-up), while the controller module receives external inputs from target sources (top-down).
Each bottom-up input is associated with spiking patterns from top-down inputs representing a target activity for the output neuron.
The controller module is only active during training and sends feedback signals proportional to the activity differences between the output neuron and the target sources.
These feedback signals drive the neuron to match the target activities during training and provide a learning signal for credit assignment on the feedforward synapses \cite{meulemansCreditAssignmentNeural2021,aceitunoLearningCorticalHierarchies2023}.
%
%
This computational unit can be directly implemented on neuromorphic devices and scaled to neural networks for on-chip programming of SNNs (Figure \ref{fig:model-binary-classification}a). 
In this scenario, the feedback signals to the output neuron are also propagated to every layer, guiding the entire network toward the target activities.

In this work, we restricted the implementation to a one-layer network.
Accordingly, the network comprises $n$ output neurons (one neuron per input class). 
The controller module comprises $n$ pairs of positive and negative control neurons associated with each output neuron.
First, the dynamics of the membrane voltage and the spiking activity of the output neurons are described by the following equations,
\begin{equation}
\begin{cases}
\mb{v}^{\text{out}}_t = \alpha \,\mb{v}^{\text{out}}_{t-1} - v_{th} \, \mb{s}^{\text{out}}_{t-1} + \mb{I}^{\text{ff}}_t + \mb{I}^{\text{fb}}_t  \\[1pt]
\mb{s}^{\text{out}}_t = H(\mb{v}^{\text{out}}_t - v_{th})\,,
\end{cases}
\label{eq:dynamics-output-neurons}
\end{equation}
where $\alpha = 1 - \Delta t / \tau_m$ is the decay factor given by the membrane time constant $\tau_m$ and the timestep size $\Delta t$, $v_{th}$ is the voltage threshold, $\mb{v}_t \in \mathbb{R}^n$ is the vector of membrane voltages of the network, $\mb{s}_t \in \mathbb{R}^n$ is the vector of binary spiking output of the network such that $s_{i,t} \in [0,1]$ for every $i$-th neuron, and $H(\cdot)$ is the Heaviside function. 
Here, $\mb{I}^{\text{ff}}_t$ and $\mb{I}^{\text{fb}}_t $ represent the synaptic currents of the  external inputs and the feedback inputs from the control module, as described by the following equations,
\begin{equation}
\begin{cases}
\mb{I}^{\text{ff}}_t = \beta \,\mb{I}^{\text{ff}}_{t-1} + \mb{W} \mb{s}^{\text{in}}_{t} \\
\mb{I}^{\text{fb}}_t = \beta \,\mb{I}^{\text{fb}}_{t-1} + \mb{Q}_p \mb{s}^p_{t-1} + \mb{Q}_n \mb{s}^n_{t-1} \,,
\end{cases}
\label{eq:currents-output-neurons}
\end{equation}
where $\beta = 1 - \Delta t / \tau_c$ is the decay factor given by the synaptic time constant $\tau_c$ and the timestep size $\Delta t$, $\mb{s}^{\text{in}}_{t} \in \mathbb{R}^m$ is the vector of spiking activity from the pre-synaptic inputs to the output neurons, $\mb{W} \in  \mathbb{R}^{n, m}$ is the set of feedforward weights from the $m$ pre-synaptic inputs to the $n$ output neurons, $\mb{s}^p_{t} \in \mathbb{R}^n$ and $\mb{s}^n_{t} \in \mathbb{R}^n$ are the vectors of spiking activities of the positive and negative control neurons, and $\mb{Q}^p \in \mathbb{R}^{n, n}$ and $\mb{Q}^n \in \mathbb{R}^{n, n}$ the respective feedback weights.
The feedforward currents $\mb{I}^{\text{ff}}_t$ accumulate over time the weighted spiking activity of the feedforward inputs with a decaying memory dictated by $\tau_c$.
Similarly, the feedback currents $\mb{I}^{\text{fb}}_t$ accumulate the weighted spiking activity of the control neurons.

Next, we modeled both control neurons as LIF neurons receiving spikes from the output neurons of the network and from the target sources (Figure \ref{fig:model-binary-classification}a).
The dynamics of the membrane voltage and the spiking activity of the control neurons are described by the following equations,
\begin{equation}
\begin{cases}
\mb{u}^p_t = \gamma\, \mb{u}^p_{t-1} - u_{th} \,\mb{s}^p_{t-1} + \mb{J}_t^{\text{fb}} - \mb{J}_t^{\text{ff}}  \\
\mb{u}^n_t = \gamma\, \mb{u}^n_{t-1} - u_{th} \,\mb{s}^n_{t-1} - \mb{J}_t^{\text{fb}} + \mb{J}_t^{\text{ff}}  \\[3pt]
\mb{s}^p_t = H(\mb{u}^p_t - u_{th}) \\
\mb{s}^n_t = H(\mb{u}^n_t - u_{th}) \,,
\end{cases}
\label{eq:dynamics-control-neurons}
\end{equation}
where $\gamma = 1 - \Delta t / \tau_u$ is the decay factor given by the membrane time constant $\tau_u$ and the timestep size $\Delta t$, $u_{th}$ is the voltage threshold of the control neurons, $\mb{u}^p_t \in \mathbb{R}^n$ and $\mb{u}^n_t \in \mathbb{R}^n$ are, respectively, the vector of membrane voltages of the positive and negative control neurons, and $\mb{s}^p_t \in \mathbb{R}^n$ and $\mb{s}^n_t \in \mathbb{R}^n$  the vector of binary spiking output of the positive and negative control neurons.
The spikes from the output neurons and target sources are accumulated into two respective synaptic current variables $\mb{J}^{\text{ff}}_t$ and $\mb{J}^{\text{fb}}_t$, as follows,
\begin{equation}
\begin{cases}
\mb{J}^{\text{ff}}_t = \beta \,\mb{J}^{\text{ff}}_{t-1} + \mb{s}^{\text{out}}_{t} \\
\mb{J}^{\text{fb}}_t = \beta \,\mb{J}^{\text{fb}}_{t-1} + \mb{s}^{\text{trg}}_{t} \,,
\end{cases}
\label{eq:currents-control-neurons}
\end{equation}
where $\beta = 1 - \Delta t / \tau_c$ is the same decay factor as in Equation \eqref{eq:currents-output-neurons}, $\mb{s}^{\text{out}}_{t} \in \mathbb{R}^n$ is the vector of spiking activity from the output neurons of the network (see Equation \eqref{eq:dynamics-output-neurons}), and $\mb{s}^{\text{trg}}_{t} \in \mathbb{R}^n$ is the vector of spiking activity from the target sources.
We refer to the next section for a detailed description of the optimization algorithm.

\subsection{Optimization with spiking feedback control}
\label{sec:methods:optimizer-learning-rule}
In this work, we use the feedback control optimizer to train SNNs on classification tasks.
We use the firing rates of the output neurons $\mb{r}^{\text{out}} \in \mathbb{R}^n$  as the readout variables for classification, 
\begin{equation}
    \mb{r}^{\text{out}} = \frac{1}{T}\sum_{t=0}^T\mb{s}^{\text{out}}_t\,,
\end{equation}
where $\mb{s}^{\text{out}}_t$ is the spiking activity from the output neurons (see Equation \eqref{eq:dynamics-output-neurons}).
We formalize a standard classification problem by defining a cross-entropy objective,
\begin{equation}
    \mathcal{L}(\hat{\mb{y}}, \mb{y}) \equiv - \sum_i \mb{y}_i\log(\hat{\mb{y}}_i) \,,
\label{eq:class-loss}
\end{equation}
where $\hat{\mb{y}}_i = \text{softmax}(\mb{r}^{\text{out}}_i) \in \mathbb{R}^n$ is the vector of predictions from the network for the $i$-th sample, and $\mb{y}_i \in \mathbb{R}^n$ the one-hot encoded true labels.
For every $i$-th sample, the true labels $\mb{y}_i$ are associated with a set of target firing rates $\mb{r}_i^{\text{trg}} \in \mathbb{R}^n$ where the correct class has a higher target activity than the incorrect classes.
%
%
We quantify the difference between target and output activities - the target error -  with a complementary objective function, 
\begin{equation}
    \mathcal{L}_{\text{trg}}(\mb{r}_i^{\text{trg}}, \mb{r}^{\text{out}}_i) = \sum_i \big| \mb{r}_i^{\text{trg}} - \mb{r}^{\text{out}}_i \big| \,,
\label{eq:target-loss}
\end{equation}
that is, the mean absolute difference between the target activity $\mb{r}^{\text{trg}}$ and the output firing rate of the network $\mb{r}^{\text{out}}$.
Since the target activities hold the same structure as the one-hot encoded labels, optimizing the network for the objective in Equation \eqref{eq:target-loss} implicitly minimizes the classification loss in Equation \eqref{eq:class-loss} \cite{meulemansCreditAssignmentNeural2021, meulemansLeastcontrolPrincipleLocal2022}.
%
%
In our framework, we optimize the network for the objective \eqref{eq:class-loss} with a feedback control approach \cite{meulemansCreditAssignmentNeural2021}.
The control system defined in Equation \eqref{eq:dynamics-control-neurons} implements a spiking PI controller that sends feedback proportional to an estimate of the difference between the target activity $\mb{r}^{\text{trg}}$ and the output activity $\mb{r}^{\text{out}}$.
Indeed, the spiking activity $\mb{s}^{\text{trg}}_{t}$ 
is modeled as an homogeneous Poisson process with rate $\mb{r}^{\text{trg}}$, such that $\mathbb{E}[\mb{s}^{\text{trg}}_{t}] = \mb{r}^{\text{trg}}\Delta t$, where $\Delta t$ is the timestep size. 
It follows from Equation \eqref{eq:currents-control-neurons} that the expected value of the current $\mb{J}^{\text{fb}}_t$ can be derived as,
\begin{equation}
\begin{split}
\lim_{t \rightarrow \infty} \,\mathbb{E}[ \mb{J}^{\text{fb}}_t] & = \lim_{t \rightarrow \infty} \mathbb{E}\Big[ \sum_{t'<t} \beta^{t-t'}\mb{s}^{\text{trg}}_{t'} + \beta^t \mb{J}^{\text{fb}}_0 \Big] \\
& = \lim_{t \rightarrow \infty} \,\Big( \sum_{t'<t} \beta^{t-t'}\mb{r}^{\text{trg}}\,\Delta t  + \beta^t \mathbb{E}[\mb{J}^{\text{fb}}_0] \,\Big)\\
& = \lim_{t \rightarrow \infty} \,\Big(\mb{r}^{\text{trg}}\,\Delta t \sum_{t'<t} \beta^{t-t'} + \beta^t \mathbb{E}[\mb{J}^{\text{fb}}_0] \Big) \\
&
=  \tau_c\,\Delta t \,\mb{r}^{\text{trg}} \,,
\end{split}
\end{equation}
which is proportional to the target firing rate $\mb{r}^{\text{trg}}$ as $t \rightarrow \infty$.
Furthermore, this current implements an exponential moving average of the underlying firing rate, with decay factor $\beta^{t-t'}$ determined by $\tau_c$ (see Equation \eqref{eq:currents-control-neurons}).
When this time constant is longer than the average inter-spike interval $1 / \mb{r}^{\text{trg}}$, then the current $\mb{J}^{\text{fb}}_{t}$ approximates the moving average of the underlying firing rate, that is, 
\begin{equation}
    \mb{J}^{\text{fb}}_{t} \appropto  \mb{r}^{\text{trg}} \,\quad\text{for} \,\, t  \gg \tau_c \,.
\end{equation}
Similarly, the input spiking activity $\mb{s}^{\text{in}}_t$ follows a homogeneous Poisson process with rate $\mb{r}^{\text{in}}$,  where $\mathbb{E}[\mb{s}^{\text{in}}_{t}] = \mb{r}^{\text{in}}\Delta t$.
The same analysis can be then applied to the synaptic current $\mb{I}^{\text{ff}}_t$ in Equation \eqref{eq:currents-output-neurons}, yielding
\begin{equation}
    \mb{I}^{\text{ff}}_{t} \appropto  \mb{r}^{\text{in}} \,\quad\text{for} \,\, t \gg \tau_c \,.
\end{equation}
Under the assumptions that (1) $\tau_c$ is larger than the average inter-spike interval of both control and output neurons, and (2) the reset dynamics are negligible due to $\tau_u < \tau_c$, the synaptic currents in Equation \eqref{eq:currents-output-neurons} and Equation \eqref{eq:currents-control-neurons} approximates the underlying firing rate as follows,
\begin{equation}
    \mb{J}^{\text{ff}}_{t} \appropto  \mb{r}^{\text{out}} \,\,\,;\,\,\, \mb{I}^{\text{fb}}_{t} \appropto  \mb{r}^{\text{p}} + \mb{r}^{\text{n}} \,,
\label{eq:feedback-current}
\end{equation}
where $\mb{r}^{\text{p}}$ and $\mb{r}^{\text{n}}$ is the firing rate of the positive and negative control, respectively.
Therefore, Equations \eqref{eq:dynamics-control-neurons} represent two spiking PI controllers driven by the estimated difference between the target and output firing rates. 
%
%

The output neurons use feedback signals from the control module to guide learning \cite{meulemansCreditAssignmentNeural2021}.
Specifically, the difference in neuronal activity with feedback $\mb{r}^{\text{fb}}$ versus without feedback $\mb{r}^{\text{ff}}$ encodes error information which drives weight updates according to
\begin{equation}
    \Delta \mb{W}\big|_t = (\mb{r}_t^{\text{fb}}
 - \mb{r}_t^{\text{ff}})\, {\mb{r}_t^{\text{in}}}^\top \,,
\end{equation}
where $\mb{r}^{\text{in}}$ is the firing rates of the pre-synaptic neurons.
The firing rate of a LIF neuron is approximately proportional to the total input current $\mb{I}_t^{\text{tot}}$ $\mb{r}_t$ as follows,
\begin{equation}
    \mb{r}_t \equiv \mb{r}_t(\mb{I}^{\text{tot}}_t) \approx  k\,\big(\mb{I}_t^{\text{tot}} - \bar{\mb{I}}\big)\quad \text{if} \quad \mb{I}_t^{\text{tot}} > \bar{\mb{I}}\,,
\end{equation}
where $k$ is a proportionality constant, $\mb{I}^{\text{tot}}_t$ is the total synaptic currents, and $\bar{\mb{I}}$ is the minimal threshold current that drives post-synaptic spikes.
Substituting this into the weight update equation, we obtain
\begin{equation}
\begin{split}
\Delta \mb{W}\big|_t & = (\mb{r}_t^{\text{fb}}
 - \mb{r}_t^{\text{ff}})\, \mb{s}_t^\top \\[3pt] 
& \approx k(\mb{I}_t^{\text{fb}} + \mb{I}_t^{\text{ff}} - \bar{\mb{I}}
 - \mb{I}_t^{\text{ff}} - \bar{\mb{I}})\, \mb{s}_t^\top \\[3pt]
& = k\,\mb{I}_t^{\text{fb}}\,\mb{s}_t^\top\,,
\end{split}
\end{equation}
and thus the following learning rule,
\begin{equation}
    \mb{W}\big|_t = \mb{W}\big|_{t-1} + \eta\,\mb{I}_t^{\text{fb}}\,\mb{s}_t^\top 
\label{eq:learning-rule}
\end{equation}
where $\eta$ is the learning rate.
From \eqref{eq:feedback-current}, it follows that weight updates are driven by and proportional to the firing activity of the control neurons.

\section{Results}
\label{sec:results}

\subsection{Classification tasks with the feedback control optimizer}
\label{sec:results:classification}
In the following, we demonstrate that the feedback control optimizer effectively trains single-layer SNN networks.
To do so, we evaluate the algorithm's performances on two different datasets (see \nameref{sec:methods:datasets}).
\begin{figure*}[!htbp]
\includegraphics[width=\textwidth]{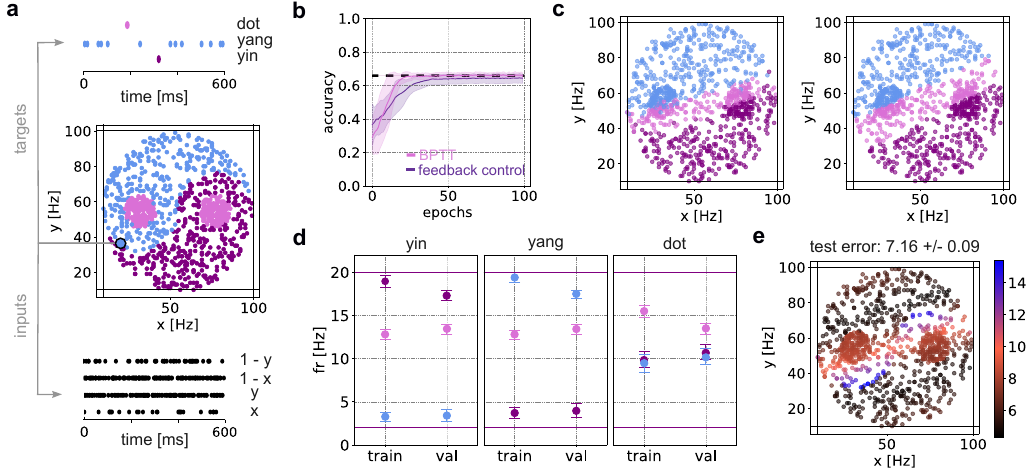}
\caption{
\textbf{Training a single-layer SNN on the spiking Yin-Yang dataset with feedback control}.
\textbf{a}) 
Illustration of the spiking Yin-Yang dataset and the corresponding targets.
Each dot belongs to the Yin (purple), Yang (azure), or dot (pink) region.
Here we illustrate the spiking patterns of the input coordinates (bottom) and target activities (top) of an example dot from the Yang class.
\textbf{b})
The average validation accuracy for the feedback control network (purple) compared to a network of three leaky-integrator neurons trained with backpropagation-through-time (BPTT).
The dashed blue line represents test accuracy for a standard readout layer trained with backpropagation, using firing rates directly as features for each dot's coordinates.
\textbf{c})
Left) The network predictions on the test set, with a test accuracy of 0.63 $\pm$ 0.03.
Right) The predictions of a standard ANN on the test set, with a test accuracy of 0.66
 $\pm$ 0.02.
\textbf{d})
The mean firing rate of the output neurons at the final epoch for both training (train) and validation (val) inputs across the Yin, Yang, and Dot classes.
Each output neuron is color-coded according to the class it represents: purple for Yin, azure for Yang, and pink for Dot.
\textbf{e})
The mean target error on the test set, with dots color-coded by associated target error (see color bar).
We calculate the mean and standard deviation across multiple simulations with different random seeds.
}
\label{fig:yin-yang-classification}
\end{figure*}
We first demonstrate the dynamics of the feedback control with a binary classification task.
In this setting, one output neuron receives spikes from two pre-synaptic inputs A (black) and B (grey) (Figure \ref{fig:model-binary-classification}b).
The inputs belong to two different classes: class 1 where input A fires at 100 Hz and input B at 50 Hz, and class 2 where input A fires at 50 Hz and input B at 100 Hz.
The neuron is trained to reach specific target activities for each class (100 Hz for class 1 and 20 Hz for class 2).
During training, the control neurons integrate the difference between the output neuron’s activity (blue) and the target spikes (purple) into their membrane potential.
When this potential reaches a threshold, the control neurons generate positive or negative feedback, adjusting the output neuron’s activity to better match the target (Figure \ref{fig:model-binary-classification}c).
During validation and testing, the control module is inactive, and the output neuron relies solely on its learned synaptic weights $W_A$ and $W_B$ to reach the target activities.
The feedback control drives the neuron to achieve zero classification loss during training and zero validation error in a few epochs, reaching  100\% accuracy during the whole training process (Figure \ref{fig:model-binary-classification}d). 
Consequently, the difference between the output and target activities without feedback control consistently decreases across epochs (Figure \ref{fig:model-binary-classification}d).
The neuron progressively reaches the target activities without feedback control for both class types when trained to encode for either class 1 or class 2 (Figure \ref{fig:model-binary-classification}e).
These results demonstrate that the control system can precisely drive the neuron activity toward target firing rates.
The output neuron changes its synaptic weights proportionally to the feedback and learns to reach the target without the feedback.
\begin{figure*}[!htbp]
\includegraphics[width=\textwidth]{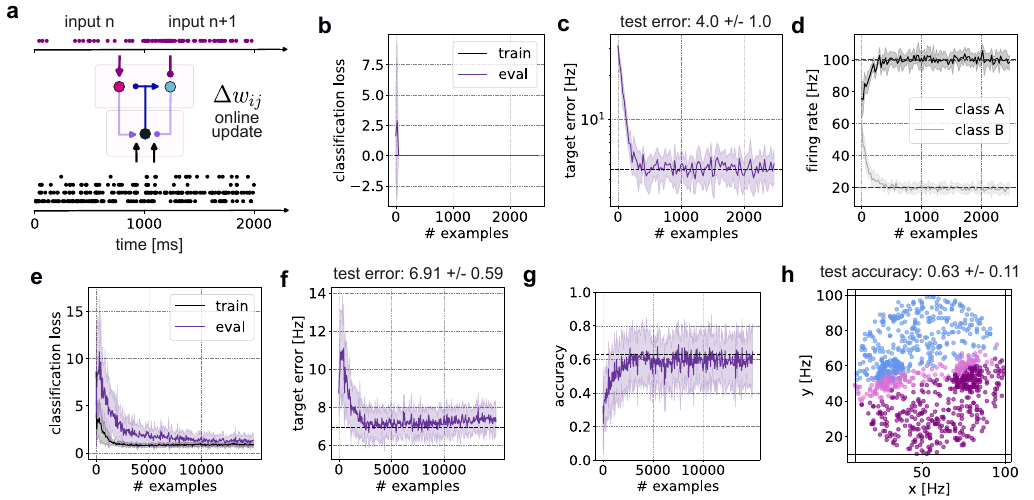}
\caption{
\textbf{Online learning in a single-layer SNN with feedback control}.
\textbf{a}) Schematic of the online learning setup.
(\textbf{b-d})
Binary classification tasks.
\textbf{b})
The average cross-entropy loss during training (black) and validation (purple). 
We calculate the mean and standard deviation by averaging over 25 sequential inputs.
\textbf{c})
Same as \textbf{b} for the mean absolute difference between output and target activity during validation.
\textbf{d})
Same as \textbf{b} for firing rate during validation for inputs from class A (black) and class B (grey) for a neuron encoding class A.
(\textbf{e}-\textbf{h})
The Spiking Yin-Yang dataset:
\textbf{e}) 
The average cross-entropy loss during training (black) and validation (purple). 
We calculate the mean and standard deviation by averaging over 50 sequential inputs.
\textbf{f})
Same as \textbf{e} for the mean absolute difference between output and target activity during validation.
The test target error is calculated offline over the whole testing set with batch size 20.
\textbf{g})
Same as \textbf{e} for the average classification accuracy during validation. 
\textbf{h})
The average network predictions when testing offline over the whole testing set with batch size 20.
We calculate the mean and standard deviation across multiple simulations with different random seeds.
}
\label{fig:online-learning}
\end{figure*}

Next, we train a one-layer SNN on a spike-based version of the Yin-Yang dataset \cite{krienerYinYangDataset2022}.
This dataset comprises dots from the Yin, Yang, and Dot regions.
The position of each dot is determined by four coordinates $x$, $y$, $1-x$, and $1-y$.
We encode these features as firing rates between 10 and 100 Hz, where each coordinate is represented by a homogeneous Poisson process with the given rate (Figure \ref{fig:yin-yang-classification}a).
We train the three output neurons to achieve a target firing rate of 20 Hz for the correct class and 2 Hz otherwise (Figure \ref{fig:yin-yang-classification}a). 
We train the network with the feedback control optimizer for 100 epochs with a batch size of 20.
We benchmark our results against (1) a single-layer network of leaky integrator units trained with backpropagation-through-time (BPTT) and (2) a linear readout layer using the 4 input firing rates as representations of each dot position.
The SNN trained with feedback control achieves a comparable accuracy to the other two models trained with backpropagation (Figure \ref{fig:yin-yang-classification}b).
Furthermore, the SNN produces similar predictions on the test set as those expected from a linear readout model \cite{krienerYinYangDataset2022} (Figure \ref{fig:yin-yang-classification}c).
Accordingly, the neurons encoding the Yin/Yang classes show a firing rate closer to the high target for inputs of its respective class and close to the low target for inputs from the opposite class (Figure \ref{fig:yin-yang-classification}d).
In contrast, the neuron encoding the Dot class shows similar firing rates across all classes, with a slight increase for Dot inputs.
We note that during training (with active control) the output neurons reach the target firing rates more effectively than during validation (without active control).
The target error is higher in the regions where the model provides wrong class predictions, in particular at the borders between the Ying and Yang regions (Figure \ref{fig:yin-yang-classification}e).

In summary, these results show that the feedback control optimizer effectively trains single-layer SNNs.
The network matches the performance of backpropagation-trained models with neurons approaching the target firing rates when the feedback is active (training) and inactive (validation).
The next step for testing the applicability of this optimizer for mixed-signal devices is to examine its performances under two important conditions: (1) online learning with a continuous input stream and (2) hyperparameter mismatch with analog components. 

\subsection{Online learning with feedback control}
In this section, we test the first condition and we simulate the performance of the models during online learning.
In this setting, the network receives inputs and targets sequentially without resetting the dynamics of both the output and the control neurons (see \textit{Online learning} section in \nameref{sec:methods:numerical-simulations}).
Moreover, the synaptic weights $w_{ij}$ are updated continuously during training examples, each time a pre-synaptic spike $s_j$ reaches the post-synaptic neuron (Figure \ref{fig:online-learning}a). 

When optimizing the network on the binary classification task, we observe that the network achieves zero classification loss during training and zero validation loss after training on the first 500 input examples (Figure \ref{fig:online-learning}b).
Similarly to our results in Figure \ref{fig:model-binary-classification}, the target error decreases during training and reaches comparable performances to the offline training (Figure \ref{fig:online-learning}c).
On average, the neuron consistently reaches the target activities (Figure \ref{fig:online-learning}d).
Furthermore, we show in Figure \ref{fig:online-learning}e that the classification and validation error on the spiking Yin-Yang dataset also decreases when training the network online.
The network reaches the same average accuracy as in the offline case after the first 50000 examples (Figure \ref{fig:online-learning}f).
Accordingly, the predictions of the network are consistent with offline training (Figure \ref{fig:online-learning}g).
The target error decreases during online training with a target error on the test set that is quantitatively comparable with offline training (Figure \ref{fig:online-learning}h).

In conclusion, these results show that the feedback control optimizer is suitable for online learning applications.
This capability arises from the single-phase structure of the learning rule in Equation \ref{eq:learning-rule}, which is driven by feedback control.
Our results indicate that embedding error information directly into synaptic currents is essential for simultaneously adjusting neuronal activity and controlling weight updates.

\subsection{Impact of device mismatch on classification performances}
\begin{figure*}[htbp]
\includegraphics[width=\textwidth]{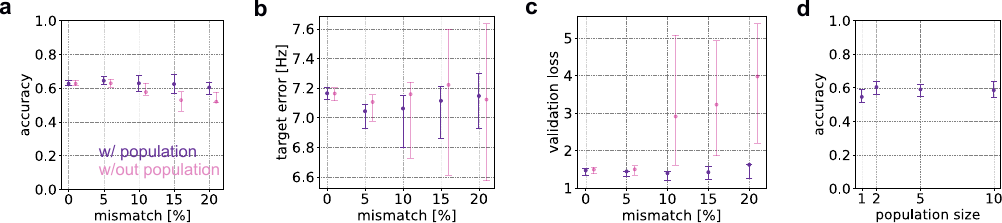}
\caption{
\textbf{Robustness of the feedback control optimizer to device mismatch}.
\textbf{a}) 
The test classification accuracy for different values of simulated device mismatch (in percentage). 
The performance are calculated with and without population averaging with $p$ = 2 (purple and pink, respectively).
The median and inter-quartile ranges are calculated over 15 simulations with different random seeds.
\textbf{b})
Same as \textbf{a} for the test target error.
\textbf{c})
Same as \textbf{a} for the validation loss.
\textbf{d})
The test classification accuracy for different population sizes with a given mismatch (20\%). 
}
\label{fig:hardware-aware}
\end{figure*}
Finally, we evaluate the second condition for hardware implementation by analyzing the performance of the feedback control optimizer under varying levels of device mismatch  (see the Methods section \nameref{sec:methods:numerical-simulations}).
We train the same network as in Figure \ref{fig:yin-yang-classification}  introducing simulated device mismatches ranging from 0\% to 20\%, and quantify how these mismatches affect model performance.
Next, we repeat the same analysis using populations of all-to-all connected neurons, simulating $p$ neurons for each output and control neuron. 
In these simulations, each neuron is independently affected by device mismatch in its model parameter.

Classification accuracy without population averaging remains stable for mismatch levels up to 10\%, with significant degradation only at higher levels (Figure \ref{fig:hardware-aware}a).
Notably, using populations of just two neurons improves test classification accuracy and restores performance to levels observed without mismatch  (Figure \ref{fig:hardware-aware}a).
Additionally, the test error shows reduced variability around the median with two-neuron populations (Figure \ref{fig:yin-yang-classification}b).
The validation loss at the end of training remains nearly constant across all tested mismatch levels, comparable to the scenario without population averaging (Figure \ref{fig:hardware-aware}c)
Increasing the population size further recovers performance despite device mismatches, with a slight decline in performance for larger populations  (Figure \ref{fig:hardware-aware}d).

Our robustness test simulations demonstrate that the feedback control optimizer maintains high classification accuracy even under significant device mismatch. 
Furthermore, introducing neural populations of $p = 2$ fully restores the original performance, aligning well with the capabilities of current neuromorphic technologies \cite{moradiScalableMulticoreArchitecture2018,richterDYNAPSE2ScalableMulticore2023}.

\section{Discussion}
\label{sec:discussion}
%
%
Efficient on-chip optimization algorithms are crucial for unlocking the full potential of mixed-signal neuromorphic devices in real-world applications. 
When paired with advancements in neuromorphic hardware, these algorithms can surpass standard machine learning methods in both power efficiency and adaptability.
Building on recent advances in biologically plausible learning \cite{meulemansCreditAssignmentNeural2021,meulemansMinimizingControlCredit2022,aceitunoLearningCorticalHierarchies2023}, we propose a feedback control optimizer specifically designed for online learning directly on hardware.
Specifically, we introduce a novel architectural primitive where a control layer drives SNNs to reach target activities, while weights are optimized with a fully local learning rule.
Our approach combines inference and learning in one single phase by relying on the recurrent connections between the network and the control, allowing for online learning on neuromorphic devices.
Here, we first validate our method on a spike-based binary classification task.
Next, we show that the feedback control optimizer can effectively train single-layer SNNs during both offline and online learning settings.
Finally, we quantify the robustness of the algorithm against typical hardware mismatches, showing minimal loss in performance.
Notably, our results show that the original performance levels can be fully restored by modestly increasing the number of neurons.
Together, our study demonstrates that our feedback control optimizer can train SNNs online while effectively compensating for typical mismatches in mixed-signal neuromorphic devices. 
This work has the potential to advance the scalability and applicability of neuromorphic hardware significantly.
%
%

Previous work has proposed to train neuromorphic devices by taking inspiration from synaptic plasticity mechanisms observed experimentally \cite{markramRegulationSynapticEfficacy1997,sjostromRateTimingCooperativity2001,bittnerBehavioralTimeScale2017}.
The close alignment between these biological principles (binary communication, eligibility traces, and voltage dynamics, among others) and the functional design of neuromorphic devices makes these rules particularly well-suited for on-chip applications.
While several spike-timing-dependent mechanisms have been successfully implemented on mixed-signal devices \cite{indiveriNeuromorphicBisableVLSI2002,chiccaVLSIRecurrentNetwork2003,mitraRealTimeClassificationComplex2009,ramakrishnanFloatingGateSynapses2011}, these models often face limitations in expressivity and typically require additional mechanisms to ensure stability \cite{rubinoNeuromorphicAnalogCircuits2023}.
%

Alternatively, it is possible to train multi-layered SNNs offline with gradient-based learning by combining BPTT \cite{rumelhartLearningRepresentationsBackpropagating1986,werbosBackpropagationTimeWhat1990} and surrogate gradients \cite{neftciSurrogateGradientLearning2019,zenkeRemarkableRobustnessSurrogate2021}.
This approach effectively bypasses the on-chip credit assignment problem by leveraging a conventional Von Neumann architecture to compute error backpropagation through automatic differentiation.
However, this approach requires storing observables from the mixed-signal hardware over time, maintaining the computational graph, waiting for the forward pass to complete, retrieving stored information to calculate weight updates offline, and then reconfiguring the on-chip weights accordingly \cite{cramerSurrogateGradientsAnalog2022}.
These requirements make this method unsuitable for online, on-chip learning scenarios, where both inference and learning are intended to occur on the device while it is running.


Looking forward, a key direction for future work is training multi-layered SNNS with the feedback control optimizer.
This is fundamental for exploiting the scalability of the algorithm, investigating its capabilities on different datasets, and demonstrating the feasibility of on-chip optimization for multi-layered SNNs.
Moreover, it is necessary to extend the testing of the algorithm with behavioral simulations and following implementation of the architecture on mixed-signal devices such as the DYNAP-SE \cite{moradiScalableMulticoreArchitecture2018,richterDYNAPSE2ScalableMulticore2023} and Brainscales \cite{pehleBrainScaleS2AcceleratedNeuromorphic2022}.
While the optimization of these devices can initially be performed with a computer-in-the-loop setup \cite{cramerSurrogateGradientsAnalog2022}, further testing of the algorithm will pave the way for designing a blueprint for a next-generation neuromorphic device.

\section{Methods}
%
%

\subsection{Datasets}
\label{sec:methods:datasets}
In this work, we consider two different datasets.
The first dataset is a custom binary classification set with 5000 training, 1000 validation, and 1000 test examples. 
Each sample consists of spiking activity from two inputs, A and B, with firing rates $f_A$ and $f_B$.
The two classes are defined by $f_A > f_B$ for class A and $f_B > f_A$ for class B.
We define a pair of firing rates $f_1$ and $f_0$ (with $f_1 > f_0$) and assign the high $f_1$ and the low $f_0$ firing rate targets to the inputs based on their class.
We simulate a homogeneous Poisson process over $T$ timesteps with timestep size $\Delta t$, where each input and target samples have dimensions $(2,T)$ and $(1,T)$, respectively. 
The second dataset we used is a spike-based version of the Yin-Yang dataset \cite{krienerYinYangDataset2022}.
Each sample in the original dataset represents a point in a 2D space encoded by coordinates $x$, $y$, $1-x$, and $1-y$, and classified into three classes (Yin, Yang, Dot) representing the yin-yang symbol. 
We preprocess this dataset by mapping each coordinate (values 0 to 1) to a firing rate with possible values $f_{min}$ and $f_{max}$.
We map the one-hot encoding of the correct class to a pair of firing rates $f_1$ and $f_0$ (with $f_1 > f_0$) and assign the high target $f_1$ and the low target $f_0$ for the correct class and the incorrect classes, respectively.
Finally, we simulate a homogenenous Poisson process for each of the four input firing rates and each of the three target firing rates.
Each input and target samples have dimensions $(4,T)$ and $(3,T)$, respectively, with 5000 training samples and 1000 each for validation and testing \cite{krienerYinYangDataset2022}.

\subsection{Numerical simulations}
\label{sec:methods:numerical-simulations}
\subsubsection{Offline training}
We train the feedback control network with the binary classification dataset for 30 epochs with a batch size of 50 during both training and evaluation,  learning rate $\eta = 10^{-5}$, and a total number of timestep $T = 5000$.
The feedforward weights $\mb{W}$ are initialized randomly from a uniform distribution between $[0, 0.04]$. 
The feedback weights $\mb{Q}_p$ and $\mb{Q}_n$ are initialized as the identity matrix $\mb{Q}_p = \mb{Q}_n = \mathbb{I}_{n,n}$, such that each output neuron receives feedback only from its respective positive and negative control.
When using the spiking Yin-Yang dataset, we train the models for 100 epochs with a batch size of 50 during both training and evaluation, learning rate $\eta = 1e10^{-4}$ for the feedback control network ($\eta = 5e10^{-4}$ for the linear leaky readout network), and a total number of timestep $T = 1000$.
When using the standard Yin-Yang dataset, we train a linear readout layer for 300 epochs with a batch size of 20 during both training and evaluation, learning rate $\eta = 2 e10^{-3}$, and input firing rates estimated from homogeneous Poisson processes with a total number of timestep $T = 1000$.
For both types of Yin-Yang datasets, the feedforward synaptic weights $\mb{W}$ are initialized randomly from a Gaussian distribution with zero mean and  $1/\sqrt{d_{in}}$ standard deviation (where $d_{in} = 4$). 
We set the timestep size $\Delta t = 1$ ms for every simulation.
We randomly initialize the training, validation, test sets, and feedforward weights with a different seed. 
We estimate the average performance metrics and network firing rates across different random seeds (5 seeds for the binary classification tasks, 15 seeds for the Ying-Yang and Spiking Yin-Yang datasets).

\subsubsection{Online learning}
We train the models with a batch size of 1 (input and target samples are selected at random and processed sequentially) without resetting the model state variables between samples.
Moreover, the synaptic weights $\mb{W}$ are updated continuously each time a pre-synaptic spike reaches a post-synaptic neuron during training (see Equation \ref{eq:learning-rule}). 
This continuous update occurs over sequences of 25 examples for the binary classification task and 50 examples for the Yin-Yang dataset.
We train the model with 2500 samples for binary classification (10000 samples for Spiking Yin-Yang), with $T = 4000$ ($T = 1000$) and learning rate $\eta = 1e10^{-9}$ ($\eta = 5e10^{-9}$).
We compute the average metrics (classification loss, target error, classification accuracy, and output firing rates) over these sequences of examples and across 15 different random seeds.
We initialize the dataset and the feedforward weights as during offline training.
\subsubsection{Device mismatch}
Device mismatches arise from variations in the electrical properties of identical components due to manufacturing imperfections~\cite{Pelgrom_etal89,Sun_etal19}.
In analog circuits, these mismatches typically result in a coefficient of variation (CV) of around 0.2~\cite{Zendrikov_etal23,Chicca_etal14}, although this value can vary depending on the circuit design and technology used.
To simulate device mismatches, we define a mismatch percentage and apply a random deviation to each model parameter.
This deviation is proportional to the ideal parameter value and the specified mismatch percentage. 
Each output and control neuron parameter is independently affected by random deviations based on the defined mismatch.

\section*{Acknowledgment}
M.S. was supported by the ETH Postdoctoral fellowship from ETH Zùrich (reference nr. 23-2 FEL-042).
C.D.L has received funding from: Bridge Fellowship founded by the Digital Society Initiative at University of Zurich (grant no.G-95017-01-12). 
G.I. was supported by the HORIZON EUROPE EIC Pathfinder Grant ELEGANCE (Grant No. 101161114) and has received funding from the Swiss State Secretariat for Education, Research and Innovation (SERI).
B.G. was supported by the Swiss National Science Foundation (CRSII5-173721, 315230 189251) and ETH project funding (24-2 ETH-032).

\section*{Authorship contributions}
Conceptualization: MS, CDL, GI, BG. Mathematical analysis: MS. Numerical simulations: MS. Writing: MS, CDL, GI, BG. Supervision: GI, BG.

\section*{Code availability}
The code to reproduce our results and all the Figures in the text is freely available at
\href{https://github.com/matteosaponati/snn-feedback-control-NICE-2025}{https://github.com/matteosaponati/snn-feedback-control-NICE-2025}.


\vspace{\fill} 
\newpage
\printbibliography

\end{document}